\documentclass[10pt,twocolumn,letterpaper]{article}

\usepackage{cvpr}              

\usepackage[dvipsnames]{xcolor}

\usepackage[accsupp]{axessibility}
\usepackage{lipsum}
\usepackage{pifont}
\usepackage{multirow}
\usepackage{algorithm}
\usepackage{algpseudocode}
\usepackage{color, colortbl}
\usepackage{setspace}
\usepackage{anyfontsize}

\definecolor{Gray}{gray}{0.9}
\definecolor{cvprblue}{rgb}{0.21,0.49,0.74}
\usepackage[pagebackref,breaklinks,colorlinks,citecolor=cvprblue]{hyperref}

\newcommand{\cmark}{\ding{51}}
\newcommand{\xmark}{\ding{55}}

\newcommand{\symprompt}{\mathbf{p}}
\newcommand{\symdomain}{X}
\newcommand{\syminput}{\mathbf{x}}
\newcommand{\symnet}{\mathcal{N}}
\newcommand{\symdist}{\mathcal{D}}
\newcommand{\symobjective}{J}
\newcommand{\symlikelihood}{L}

\newcommand{\track}[1]{#1}                   

\def\paperID{9247} 
\def\confName{CVPR}
\def\confYear{2024}

\title{A2XP: Towards Private Domain Generalization}

\author{Geunhyeok Yu \quad Hyoseok Hwang\\
Department of Software Convergence, Kyung Hee University, Republic of Korea\\
{\tt\small \{geunhyeok, hyoseok\}@khu.ac.kr}
}

\begin{document}
\maketitle

\begin{abstract}
Deep Neural Networks (DNNs) have become pivotal in various fields, especially in computer vision, outperforming previous methodologies. 
A critical challenge in their deployment is the bias inherent in data across different domains, such as image style and environmental conditions, leading to domain gaps. 
This necessitates techniques for learning general representations from biased training data, known as domain generalization. 
This paper presents Attend to eXpert Prompts (A2XP), a novel approach for domain generalization that preserves the privacy and integrity of the network architecture. A2XP consists of two phases: Expert Adaptation and Domain Generalization. 
In the first phase, prompts for each source domain are optimized to guide the model towards the optimal direction. 
In the second phase, two embedder networks are trained to effectively amalgamate these expert prompts, aiming for an optimal output. 
Our extensive experiments demonstrate that A2XP achieves state-of-the-art results over existing non-private domain generalization methods.
The experimental results validate that the proposed approach not only tackles the domain generalization challenge in DNNs but also offers a privacy-preserving, efficient solution to the broader field of computer vision.
\track{Code is available at \url{https://github.com/AIRLABkhu/A2XP}.}
\end{abstract}
\vspace{-5mm}

\section{Introduction}
\label{sec:intro}
Deep Neural Networks (DNNs) are recognized as the most potent models in machine learning. 
They have achieved remarkable success in various fields, particularly in computer vision, where they have surpassed previous methodologies. 
Although DNNs are versatile and universal function approximators, the data they process often carry biases related to factors such as image style~\cite{vlcs}, sensor parameters~\cite{office_home}, 
as well as painting styles~\cite{pacs}. 
These biases create distinct distributions, known as domains, with inherent gaps between them. 
The inability of DNNs to generalize across these domains necessitates an impractically large amount of unbiased training data to mitigate the model's bias.
Consequently, this limitation underscores the importance of developing techniques that can learn general representations from biased training data. 
This challenge, extensively studied in various researches~\cite{survey1,survey2}, is referred to as domain generalization.

\begin{figure}[t!]
    \centering \fontsize{6}{7}
    \def\svgwidth{1.17\linewidth} \setlength\fboxrule{0pt}
    \fbox{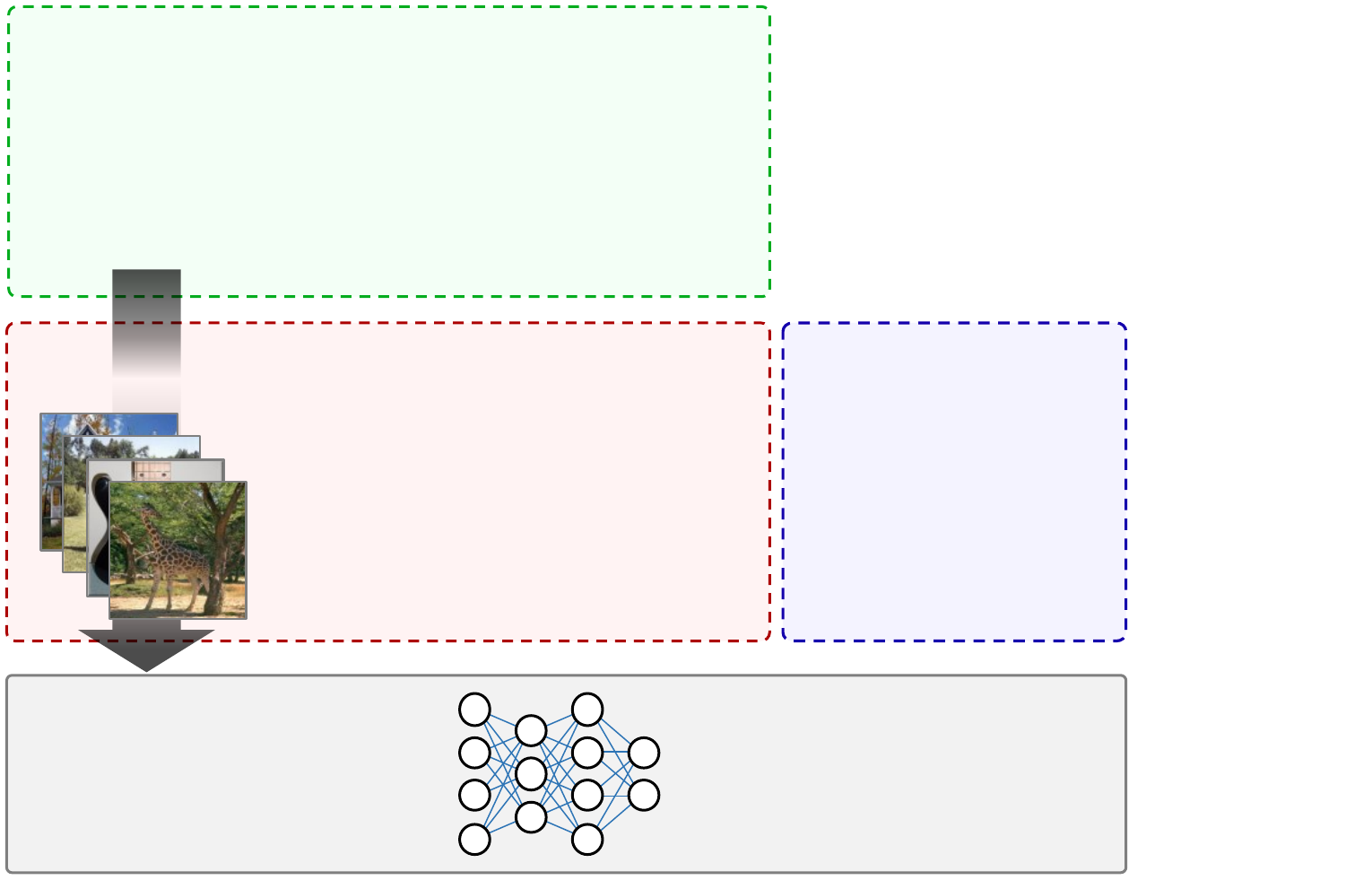}
    \caption{Flow diagram of the proposed method A2XP.}
    \vspace{-6mm}
    \label{fig:overview}
\end{figure}
To address the generalization issue, various research topics such as domain adaptation~\cite{adaptation_01,adaptation_02,adaptation_03,adaptation_04}, meta-learning~\cite{meta_learning_01,meta_learning_02,meta_learning_03}, and transfer learning~\cite{transfer_01,transfer_02,clip} have been explored. 
Domain adaptation, which shares similarities with domain generalization, specifically aims to mitigate domain gaps. 
The primary difference between the two lies in the visibility of the target domain~\cite{survey1}. 
In domain adaptation, the target domain is known and the goal is to adapt a pre-trained network to this specific domain. 
This involves learning new knowledge from the target domain while utilizing existing knowledge from source domains~\cite{adaptation_survey}, a task that is generally more straightforward than domain generalization.
In contrast, domain generalization operates without the need for target domain data, focusing on making the network robust to shift from a source domain to an unknown target domain. 
While these two approaches are distinct, the ability of domain adaptation to understand domain shifts can be beneficial for domain generalization. 
Our approach is based on this concept. 
We propose that if a network can effectively map input from any arbitrary domain into a generalized manifold space, the challenge of domain generalization could be transformed into a regression problem. 
In this scenario, adaptation strategies could provide crucial insights for determining the direction of this regression. 

Most of the methods that mitigate domain gaps necessitate access to the architecture and parameters of the target network~\cite{dann,sagnet, miro, dart, common_specific}. 
For instance, Domain Adversarial Neural Network (DANN)~\cite{dann} and Style-Agnostic Network (SagNet)~\cite{sagnet} aims to fine-tune the backbone network to extract domain-agnostic features.  Similarly, Common and Specific Visual Prompt Tuning (CSVPT)~\cite{common_specific} employs prompt tokens in conjunction with a Vision Transformer (ViT)~\cite{vit} to address these challenges. 
However, these approaches often require modifications to the network's architecture or parameters, which can pose significant privacy concerns. 

Visual Prompting~(VP)~\cite{vp} provides a solution to privacy concerns by fine-tuning an objective network through adversarial reprogramming without altering the network's architecture or parameters. 
It only tunes additional parameters known as prompts, which are added to the input image rather than being embedded within the network. 
Inspired by this, we added a prompt to the input to address the privacy issue~\cite{privacy}.
However, VP faces a limitation: an excessive number of pixels in a prompt can disrupt training. 
To overcome this, we train multiple prompts, referred to as ``experts,'' and integrate them using an attention mechanism. 
This strategy aligns with the concept of addressing domain generalization as a direction regression problem, where these experts serve as guides to identify the optimal direction for generalization.

In this study, we aim to disentangle the domain generalization problem into two steps: expert adaptation and domain generalization, while keeping the privacy of the objective network. 
We propose \textit{\textbf{A}ttend \textbf{to} e\textbf{X}pert \textbf{P}rompts (A2XP)} which is a novel domain generalization method that solves this issue.
In the expert adaptation step, we optimize prompts for each source domain to prepare the hints to find the optimal direction. 
In the domain generalization step, two embedder networks are trained to properly mix the expert prompts so that the output is in the optimal direction. 
The main contributions of this study can be summarized as follows:
\begin{itemize}
    \item{Inspired by VP, we introduce A2XP, which is a novel and simple domain generalization method that protects privacy.}
    \item{We mathematically analyze the generalization issue as an optimization of a linear combination problem.}
    \item{We further demonstrate the effectiveness and characteristics of A2XP and its component factors through extensive experiments.}
    \item{A2XP achieves SOTA over existing non-private domain generalization methods with significantly lower computational resource requirements.}
\end{itemize}

\section{Related Works}
\label{sec:related-works}

\subsection{Domain Generalization}

The objective of domain generalization is to reduce the gaps between visible source domains and unseen target domains. 
There are several approaches such as domain alignment~\cite{domain_alignment_01,miro,domain_alignment_03,domain_alignment_04,dann,domain_alignment_06,sagnet,domain_alignment_08}, meta learning~\cite{meta_learning_01,meta_learning_02,meta_learning_03,meta_learning_04}, ensemble learning~\cite{ensemble_learning_01,dart,ensemble_learning_03,ensemble_learning_04} and, representation disentanglement~\cite{meta_learning_01,disentanglement_02} as categorized by Zhou~\etal~\cite{survey1}.

Ganin~\etal~\cite{dann} introduced DANN that discriminates the domains so that the network can find domain-agnostic features. 
SagNet~\cite{sagnet} also discriminates the domains by adversarially learning content bias and style bias.
Cha~\etal~\cite{miro} aligned domains by employing a regularization term to the loss function based on mutual information among domains. 
Diversify-Aggregate-Repeat Training~(DART)~\cite{dart} is an ensemble learning method that diversifies the source domain by applying data augmentation to independently capture diverse features using multiple networks, then aggregates networks and repeats these procedures. 
DART can enhance the generalization performance, but it also takes a massive amount of memory. 

Our approach basically follows the idea of domain alignment and ensemble learning. 
We train multiple expert prompts that align source domains each. 
Then, it aggregates the experts to align a novel target domain.
The experts give a hint to find the direction to the optima of a target domain on the fly, and we take different simple generalization steps to each sample of the target domain.

\subsection{Prompt Tuning in Computer Vision}

Prompt tuning is a transfer learning technique that requires a tiny amount of additional parameters. 
Prompt tuning in computer vision was first introduced by Visual Prompt Tuning~(VPT)~\cite{vpt} for transfer learning with a small number of parameters.
VPT proved that prompt tuning is a stronger transfer learning technique than full fine-tuning and linear probing.
However, access to change the architecture of the network is required to apply VPT.
Bahng~\etal~\cite{vp} introduced adversarial reprogramming~\cite{adversarial_reprogramming}-based prompting for general pre-training using vision-language relationships.
They successfully incorporated visual and lingual representations only using an optimized perturbation to the inputs.
We will call this prompting ``input prompting''.
Huang~\etal~\cite{dam_vp} proposed Diversity-Aware Meta Visual Prompting~(DAM-VP) that transfers a network to another target dataset that contains diverse representation distribution.
DAM-VP separates a set of data into clusters and updates the prompt using each of the clusters. 
Then, it gathers all prompts from clusters to capture the diversity and provides a detailed representation of the whole data distribution.
Inspired by DAM-VP, we captured the diversity of data distribution from the source domains and generalized the target domain. 

\subsection{Attention Mechanism}

The key idea of the attention mechanism is activating important features and silencing less important features. 
Many of the modern deep learning architectures have employed attention mechanism~\cite{senet, transformer}. 
Squeeze-and-Excitation Networks~\cite{senet} focused on weighting each channel of a large feature map before aggregating them.
Transformer~\cite{transformer, vit}, one of the most effective architectures, lies its core on the attention mechanism. 
Transformers have two different types of attention mechanisms with different origins of the ``query''. 
Cross-attention builds ``query'' from the same source of ``key'' and ``value'' while self-attention builds from a different source. 
Cross-attention is used to capture the importance of ``values'' depending on the relationship with other data.
We used the cross-attention mechanism to properly combine multiple experts.

\section{Methods}
\label{sec:methods}

Domain generalization is a task that generally fits a model to unseen target domains using known source domains.
In this section, we describe A2XP, our novel domain generalization method, using input prompting. 

\begin{figure*}[t!]
    \centering \fontsize{6}{7}
    \def\svgwidth{1.1\linewidth} \setlength\fboxrule{0pt}
    \fbox{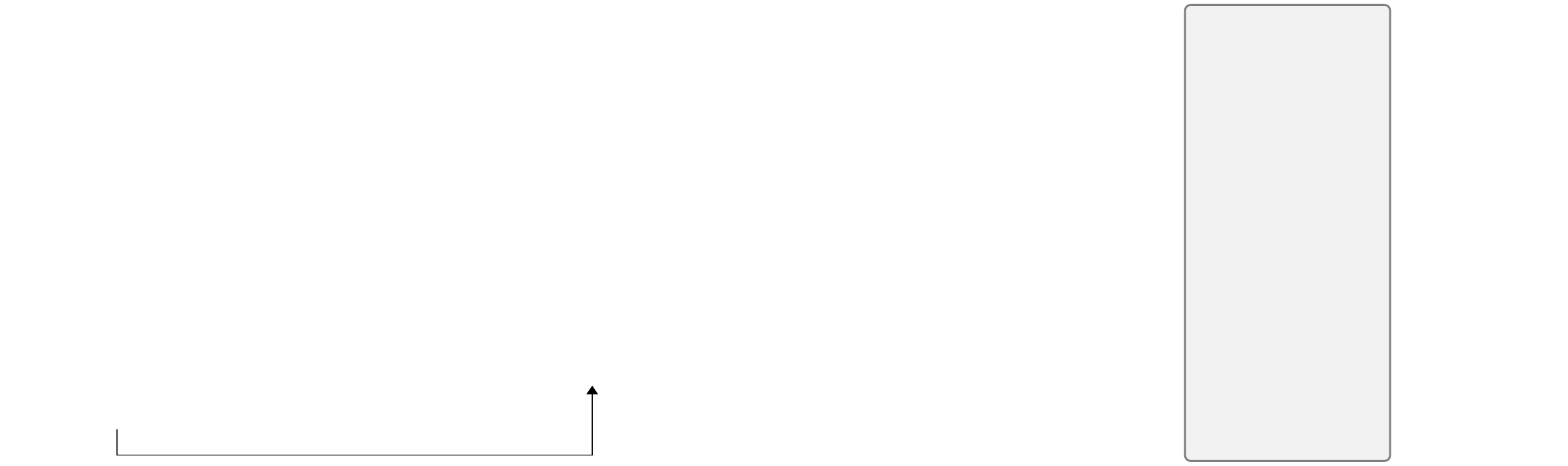}
    \vspace{-4mm}
    \caption{Inference procedure of A2XP. There are experts from source domains and target images of an unseen target domain. The experts are image-dependently mixed through an attention-based algorithm and added to the specific image.}
    \vspace{-2mm}
    \label{fig:algorithm}
\end{figure*}

\subsection{Algorithm Overview}
A2XP operates through a two-phase approach.
Initially, it performs source-wise adaptation by crafting `experts' - specific adaptation prompts for each source domain. 
This step is conducted end-to-end, predominantly via error backpropagation. 
The subsequent phase is dedicated to domain generalization, where image-specific prompts for the target domain are created for each input image by averaging the weights of all experts, determined through an attention-based algorithm. 
In this phase, the system utilizes two separate trainable encoders: one for the input images and another for the pre-trained experts.
An expert's weight is derived from the similarity between the encoded input image and the expert's embedding. 
These phases are termed \textit{Expert Adaptation} and \textit{Attention-based Generalization}, respectively. 
The A2XP algorithm is detailed in Algorithm~\ref{alg:a2xp}, with the validation process illustrated in Figure~\ref{fig:algorithm}.

\begin{algorithm}[t]
    \small \setstretch{0.9413}
    \caption{Training and Inference Scenario of A2XP} 
    \label{alg:a2xp}
    \textbf{Input:} $\symdomain_1, \symdomain_2, \cdots, \symdomain_{N+1}$ \\
    \textbf{Parameter:} Objective network $\symnet$ \\
    \textbf{Parameter:} Meta prompt $\symprompt_{\textnormal{meta}}$ \\
    \textbf{Parameter:} Learning rates $\alpha_A, \alpha_G$\\
    \textbf{Output:} Experts $\symprompt_1, \symprompt_2, \cdots, \symprompt_N$ \\
    \textbf{Output:} Encoder head parameters $\theta_{\mathcal{E}_\textnormal{T}}, \theta_{\mathcal{E}_\textnormal{E}}$ 
    \newcommand{\AlgSection}[1]{\State{\color{Green}{\leavevmode\leaders\hrule depth-2.1pt height 2.75pt\hfill\kern0pt{~$\vartriangleright$~{#1}}}}}
    
    \begin{algorithmic}[1]
        \AlgSection{Training $\symprompt_{i \in [1,N]}$}
        \For{$\symdomain_{i \in [1,N]}$}
            \State $\symprompt_i \gets \symprompt_{\textnormal{meta}}$
            \For{$(\syminput_{i,j}, \mathbf{y}_{i,j}) \in \symdomain_i$}
                \State $\symprompt_i \gets \symprompt_i - \alpha_A \partial \mathcal{L}_\textnormal{KL}(\symnet(\syminput_{i,j} + \symprompt_i), \mathbf{y}_{i,j}) / \partial \symprompt_i$
            \EndFor
        \EndFor
        \State $\symprompt_i \gets \symprompt_i / \Vert \symprompt_i \Vert_2$
        \Comment{Normalizing expert prompts}
        
        \AlgSection{Training $\theta_{\mathcal{E}_\textnormal{T}}, \theta_{\mathcal{E}_\textnormal{E}}$}
        \For{$\symdomain_{i \in [1,N]}$}
            \For{$(\syminput_{i,j}, \mathbf{y}_{i,j}) \in \symdomain_i$}
                \State $Q, K \gets \mathcal{E}_\textnormal{T}(\syminput_{i,j}), \mathcal{E}_\textnormal{E}(\symprompt_{k \in [1,N]})$
                \State $\symprompt_{i,j} \gets \sum_{k=1}^{N}{\symprompt_k \tanh(QK_k^\top)}$
                \State $l \gets \nabla \mathcal{L}_\textnormal{KL}(\symnet(\syminput_{i,j} + \symprompt_{i,j}), \mathbf{y}_{i,j})$
                \State $\theta_{\mathcal{E}_\textnormal{T}} \gets \theta_{\mathcal{E}_\textnormal{T}} - \alpha_G \partial l / \partial \theta_{\mathcal{E}_\textnormal{T}}$
                \Comment{Update $\theta$, not $\symprompt$}
                \State $\theta_{\mathcal{E}_\textnormal{E}} \gets \theta_{\mathcal{E}_\textnormal{E}} - \alpha_G \partial l / \partial \theta_{\mathcal{E}_\textnormal{E}}$
            \EndFor
        \EndFor
        
        \AlgSection{Inference on unseen $\symdomain_{N+1}$}
        \For{$\syminput_{N+1,j} \in \symdomain_{N+1}$}
            \State $Q, K \gets \mathcal{E}_\textnormal{T}(\syminput_{N+1,j}), \mathcal{E}_\textnormal{E}(\symprompt_{k \in [1,N]})$
            \State $\symprompt_{N+1,j} \gets \sum_{k=1}^{N}{\symprompt_k \tanh(QK_k^\top)}$
            \State $\hat{\mathbf{y}}_{N+1,j} \gets \symnet(\syminput_{N+1,j} + \symprompt_{N+1,j})$
            \Comment{Prediction}
        \EndFor
    \end{algorithmic}
\end{algorithm}

\subsection{Idea Formulation}

We first formulate our idea as a concrete guideline for detailed understanding.
Domain generalization using input prompting can be formulated as follows. 
For $N+1$ domains $\symdomain_{i\in[1,N+1]}$, we can select $\symdomain_{N+1}$ as a target domain and others as source domains. 
The network named $\symnet$ is given with fixed pre-trained parameters; there exist decision boundaries of the network. 
Let an expert for the $i$-th domain be $\symprompt_i \in \mathbb{R}^{d_\text{prompt}}$ where $d_\text{prompt}$ is the dimension of a prompt. 
Then, $\symprompt_{i \in [1,N]}$ represents the optimal direction that shifts the inputs in source domains and we can optimize those with the known source data.
Prompt for the target domain $\symprompt_{N+1}$ cannot be directly optimized because the target domain is invisible. 

\begin{figure}
    \vspace{-3mm}
    \centering
    \def\svgwidth{1.65\linewidth} \setlength\fboxrule{0pt}
    \fbox{\hspace{1mm} 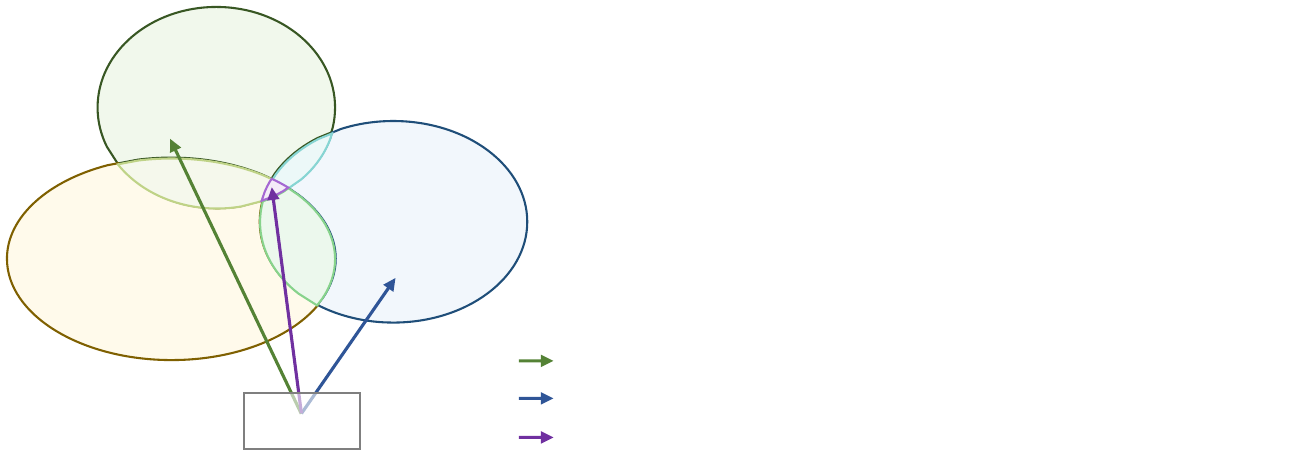}
    \vspace{-6mm}
    \caption{Geometric concept of A2XP as a linear combination in 2D manifold space with two source domains.}
    \vspace{-5.1mm}
    \label{fig:concept}
\end{figure}

We approximate $\symprompt_{N+1}$ as a linear combination of $\symprompt_{i \in [1,N]}$ as following equation:
\begin{equation}
    \label{eq:prompt_linear_combination}
    \symprompt_{N+1} = \sum_{i=1}^{N}{\lambda_i \symprompt_i}, \quad\lambda_i = \Lambda(\symprompt_i \vert \syminput \in \symdomain_i)
\end{equation}
where $\Lambda$ is a conditional function that represents the optimal weights for $\symprompt_i$ when $\syminput \in \symdomain_i$ is given. 
Let say 
\begin{equation}
    \label{eq:objective_function}
    \symobjective(\lambda_i)=\operatorname{KL}(\symnet(\syminput_{N+1} + \symprompt_{N+1}) \Vert \symdist_{N+1})
\end{equation}
be the objective function where $\syminput_{N+1} \in \symdomain_{N+1}$, $\symdist_{N+1}$ is the target distribution for $\symnet$ of $\syminput_{N+1} + \symprompt_{N+1}$ and $\operatorname{KL}$ refers to the KL-Divergence function.
Then the likelihood function $L$ has a relationship as following
\begin{equation}
    \label{eq:likelihood_function}
    \symlikelihood(\symdist_{N+1} \vert \symnet(\syminput_{N+1} + \symprompt_{N+1})) \propto
    e^{-\symobjective(\lambda_i)}.
\end{equation}
This formulation shows that
\begin{equation}
    \label{eq:negaive_log_likelihood}
    \symobjective(\lambda_i) \propto -\log \symlikelihood(\symdist_{N+1} \vert \symnet(\syminput_{N+1} + \symprompt_{N+1})),
\end{equation}
minimizing $\symobjective$ by training $\Lambda$ is equivalent to maximizing $\symlikelihood$.
This idea can be explained as follows. 
If there are ranges of optimal prompts for each domain, an expert must be a point inside the range. 
And because the target prompts are formulated as Equation~\ref{eq:prompt_linear_combination}, the geometry of the prompt space can be conceptually visualized like Figure~\ref{fig:concept}.

\subsection{Expert Adaptation}

Our objective is to mix multiple expert prompts into a single prompt.
For this to be effective, each expert must be proficiently trained in their primary field, which in our case is the domain. 
We utilize adversarial reprogramming~\cite{adversarial_reprogramming}, a straightforward gradient-based method, for adapting these experts. 
While this approach suffices in specific scenarios, it falls short in domains vastly different from the pre-training domain. 
To address this, we employed meta prompts~\cite{dam_vp} to initialize the expert prompts.
Meta prompt refers to pre-trained prompts that can be used to initialize a visual prompt. 

\subsection{Attention-based Generalization}

Our key idea is to combine the experts in a way that makes images from unseen domains to be correctly classified.
We combined experts by weight-averaging them. 
A weight must indicate how much an expert is needed for a given specific image. 
This requirement can be implemented using the cross-attention mechanism. 
In this case, the experts become ``keys'' $(K)$ and ``values'' $(V)$, a target image becomes ``query'' $(Q)$ of attention. 
The attention weight is calculated as the similarity between $Q$ and $K$. 
Instead of directly comparing $Q$ and $K$, we used embedding vectors. 
We have a pre-trained network as a shared embedder network and two different trainable head linear layers for $Q$ and $K$ each.
$Q$ and $K$ are embedded through a shared encoder and their respective linear heads.
Then scalar attention weights are obtained as much as the number of the experts, and $VQK^\top$ becomes the prompt for the target image. 
    
\begin{table*}[ht!]
    \begin{subtable}[h]{\textwidth}
        \centering \footnotesize
        \begin{tabular}{lc|ccccc|ccccc}
        \toprule
        \multirow{2}{*}{Method}  & DART~\cite{dart}         & \multicolumn{5}{c|}{PACS~\cite{pacs}}       & \multicolumn{5}{c}{VLCS~\cite{vlcs}}         \\
        & Supported & Picture & Art            & Cartoon & Sketch & Avg.  & VOC 2007 & LabelMe & Caltech101 & SUN09 & Avg.  \\
        \midrule
        SAM~\cite{sam} & \cmark    & 18.41   & 15.13          & 21.38   & 19.12  & 18.51 & 44.72    & 46.02   & 61.13      & 41.62 & 48.38 \\
        ERM~\cite{erm} & \cmark    & 97.08   & 87.19          & 86.25   & 82.38  & 88.22 & 75.60    & 64.47   & 97.08      & 77.49 & 78.66 \\
        SagNet~\cite{sagnet} & \cmark & 91.99   & 84.56          & 69.19   & 20.07  & 66.45 & 51.02    & 62.63   & 61.13      & 61.16 & 58.98 \\
        DANN~\cite{dann} & \cmark   & 97.68   & 89.93          & 86.41   & 81.11  & 88.78 & 77.86    & 66.97   & 98.59      & 73.53 & 79.24 \\
        MIRO~\cite{miro} & \cmark   & 96.48   & 90.79          & 90.46   & 83.59  & 90.33 & 78.05    & 66.68   & 97.53      & 71.97 & 78.56 \\
        \rowcolor{Gray}
        A2XP (ours) & \xmark     & \textbf{99.07}   & \textbf{95.27}        & \textbf{98.07}   & \textbf{87.85}  & \textbf{95.07} & \textbf{84.07}    & \textbf{68.72}   & \textbf{99.62}      & \textbf{80.19} & \textbf{83.15} \\
        \bottomrule
        \end{tabular}

        \caption{Comparison with other methods in the target domain. DART~\cite{dart} was applied to the baselines for their best performance.}
        \label{tab:exp_comparison_with_others_a}
        \vspace{2mm}
    \end{subtable}
    \begin{subtable}[h]{\linewidth}
        \centering \footnotesize
        \begin{tabular}{cccccc|cccccc}
        \toprule
         Source & \multicolumn{5}{c|}{Target} & Source & \multicolumn{5}{c}{Target}         \\
          & Picture & Art   & Cartoon & Sketch & \textbf{Avg.} & & VOC 2007 & LabelMe   & Caltech101 & SUN09 & \textbf{Avg.} \\
        \midrule
        P & -       & 99.88 & 99.76   & 99.52  & \textbf{99.72} & V & -        & 78.20      & 99.79      & 87.84 & \textbf{88.61} \\
        A & 96.53   & -     & 96.39   & 94.87  & \textbf{95.93} & L & 89.28    & -          & 99.36      & 84.19 & \textbf{90.94} \\
        C & 98.63   & 98.76 & -       & 98.17  & \textbf{98.52} & C & 88.48    & 78.58      & -          & 84.16 & \textbf{83.74} \\
        S & 91.45   & 91.12 & 91.98   & -      & \textbf{91.52} & S & 90.23    & 76.84      & 100.00     & -     & \textbf{89.02} \\
        \textbf{Avg.}    & \textbf{95.54}   & \textbf{96.59} & \textbf{96.04}   & \textbf{97.52}  & \textbf{96.42} & \textbf{Avg.} & \textbf{89.33} & \textbf{77.87} & \textbf{99.72} & \textbf{85.40}  & \textbf{88.08} \\
        \bottomrule

        \end{tabular}
        \caption{Source domain evaluation on PACS~\cite{pacs} (left) and VLCS~\cite{vlcs} (right) datasets.}
        \label{tab:exp_comparison_with_others_b}
    \end{subtable}

    \caption{Target domain and source domain evaluations. Target domain evaluation was conducted to compare A2XP with other state-of-the-art methods. Source domain evaluation was conducted to see if it is still effective in the source domains.}
    \vspace{-3mm}
    \label{tab:exp_comparison_with_others}
\end{table*}

However, there are two problems. 
First, the experts are independently optimized in different domains, which makes a significant difference in scales. 
We solved this by dividing the experts with the $L_2$-norm of each of themselves for normalization. 
The second problem is that the weights can be saturated too much because the weights are independently calculated without scaling such as $\operatorname{softmax}$ function. 
Mapping the weights into $[-1, 1]$ using $\tanh$ function mitigates this problem. 
As a result, the prompt $(\symprompt_{N+1,k})$ for a $k$-th target image $(\syminput_{N+1,k} \in \symdomain_{N+1})$ can be formulated as:
\begin{equation}
    \label{eq:genralization_prompt}
    \symprompt_{N+1,k} = \sum_{i=1}^{N} \frac{\symprompt_i}{\Vert \symprompt_i \Vert_2} \mathcal{E}_{\textnormal{T}}(\syminput_{N+1,k}) \mathcal{E}_{\textnormal{E}}(\frac{\symprompt_i}{\Vert \symprompt_i \Vert_2})^\top,
\end{equation}
where $\mathcal{E}_{\textnormal{T}}$ and $\mathcal{E}_{\textnormal{E}}$ denote the embedders for target images and experts respectively.
Once the generalization is trained, the embedding vectors of the experts are fixed because the experts will not be changed. 
Thus, the expert embedding procedure is no longer needed in evaluation. 

\section{Experiments and Analysis}
\label{sec:experiments}

In this section, we perform leave-one-domain-out evaluation and more extensive experiments mainly on PACS~\cite{pacs} and VLCS~\cite{vlcs} datasets and partially on Office-Home~\cite{office_home} dataset to demonstrate the effectiveness and characteristics of A2XP. 
PACS dataset consists four domains: \textbf{P}icture, \textbf{A}rt painting~(\textbf{A}rt), \textbf{C}artoon, and \textbf{S}ketch.
VLCS dataset is composed of four subdatasets, each representing a different domain:  \textbf{V}OC 2007~\cite{voc_2007}, \textbf{L}abel Me~\cite{labelme}, \textbf{C}altech101~\cite{caltech101}, and \textbf{S}UN09~\cite{sun09}.
The Office-Home~\cite{office_home} dataset consists of four domains: Art painting, Clipart, Product, and Real image. 
The experiments were conducted on Ubuntu Server 18.04 with an Intel Xeon Gold 6226R 2.90GHz and NVIDIA RTX 3090.

\subsection{Implementation Details}

For our study, we selected a CLIP~\cite{clip}-pre-trained ViT~\cite{vit} as the objective network. 
The experts within this framework were optimized through end-to-end backpropagation.
The prompt size was chosen based on the specifications of VP~\cite{vp}, which employs a padding size of 30.
We used a learning rate of 1.0E-4 and stochastic gradient descent with momentum~\cite{sgd_momentum} for optimization.
Given that a tiny network suffices for the shared embedder networks of A2XP, we opted for an ImageNet~\cite{imagenet}-pre-trained ResNet18~\cite{resnet} as the backbone.
Following the shared encoder, two distinct trainable linear heads are attached to specialize the features into $Q$ and $K$.
To demonstrate A2XP's efficiency in simplifying problems, we limited the number of updates to 1,000, unless otherwise specified. 
For optimization during generalization, we used AdamW~\cite{adamw}. 
We implemented a learning rate decay to 10\% of its initial value, utilizing the Cosine Annealing with Warm Restarts~\cite{cosineannealingwarmrestarts} algorithm, across the entire generalization procedure.

\begin{table*}[ht!]
    \centering
    \footnotesize
    \newcommand{\atob}[2]{\track{${#1}~\rightarrow~{#2}$}}
    
    \begin{tabular}{c|ccccc|ccccc}
        \toprule
        & \multicolumn{5}{c|}{Expert Adaptation} & \multicolumn{5}{c}{Attention-based Generalization} \\
        \midrule \midrule
        & \atob{P}{P} & \atob{A}{A} & \atob{C}{C} & \atob{S}{S} & Avg. & \atob{ACS}{P} & \atob{PCS}{A} & \atob{PAS}{C} & \atob{PAC}{S} & Avg. \\
        \midrule
        Zero & \textbf{97.54} & 73.88 & \textbf{95.52} & 94.55 & 90.37 & 99.07 & 95.07 & 98.12 & \textbf{88.22} & \textbf{95.12} \\
        Uniform & 78.62 & 60.25 & 87.63 & 87.76 & 78.57 & \textbf{99.15} & 94.97 & 98.17 & 88.02 & 95.08 \\
        Normal & 87.72 & 73.00 & 84.90 & \textbf{97.89} & 85.88 & 98.99 & 95.15 & \textbf{98.39} & 87.81 & 95.08 \\
        \rowcolor{Gray}
        Meta~\cite{dam_vp} & 94.07 & \textbf{93.12} & 93.60 & 93.28 & \textbf{93.52} & 99.07 & \textbf{95.27} & 98.07 & 87.85 & 95.07 \\
        \midrule \midrule
        & \atob{A}{A} & \atob{C}{C} & \atob{P}{P} & \atob{R}{R} & Avg. & \atob{CPR}{A} & \atob{APR}{C} & \atob{ACR}{P} & \atob{ACP}{R} & Avg. \\
        \midrule
        Zero & 21.18 & 38.95 & 61.93 & 43.10 & 41.29 & 67.57 & 57.98 & 66.55 & 71.29 & 65.85 \\
        Uniform & 21.63 & 32.94 & 44.92 & 46.71 & 36.55 & 67.41 & 58.33 & 67.27 & 71.77 & 66.20 \\
        Normal & 28.92 & 32.51 & 40.17 & 23.36 & 31.24 & 67.74 & 58.35 & 67.83 & 71.22 & 66.29 \\
        \rowcolor{Gray}
        Meta~\cite{dam_vp} & \textbf{47.05} & \textbf{54.39} & \textbf{69.66} & \textbf{52.03} & \textbf{56.35} & \textbf{77.42} & \textbf{65.73} & \textbf{81.93} & \textbf{83.15} & \textbf{77.06} \\
        \bottomrule

    \end{tabular}
    \caption{Generalization and adaptation performance in PACS~\cite{pacs} (top) and Office-Home~\cite{office_home} (bottom) datasets using different prompt initialization before adaptation. Zero initializes as zero tensor, Uniform initializes using uniform distribution~$\mathcal{U}(-0.03, 0.03)$, and Normal initializes using Gaussian distribution~$\mathcal{N}(0, 0.03^2)$.}
    \vspace{-3mm}
    \label{tab:exp_different_inits}
\end{table*}

\subsection{Leave-One-Domain-Out Evaluation}

We conducted a leave-one-domain-out evaluation to assess the domain generalization performance, the results of which are detailed in Table~\ref{tab:exp_comparison_with_others_a}.
In this experiment, we evaluated several methods, including domain generalization methods such as SagNet~\cite{sagnet}, DANN~\cite{dann}, and Mutual Information Regularization with Oracle (MIRO)~\cite{miro}, as well as non-domain generalization methods like Sharpness-Aware Minimization (SAM)~\cite{sam} and Empirical Risk Minimization (ERM)~\cite{erm}, following the approach used by DART~\cite{dart}.
These five baselines were augmented using DART, which is an ensemble learning-based method for domain generalization. 
A2XP outperformed all other methods in each target domain on both PACS and VLCS datasets. 
Notably, it achieved a 4.74\% increase in average accuracy on PACS dataset and a 4.99\% increase on VLCS dataset. 
It is important to mention that DART does not ensure the privacy of the objective network.

\subsection{Evaluation on Source Domains}

Domain generalization focuses on adapting models to both unseen and known source domains. 
We evaluated the generalizability of A2XP in source domains, utilizing the expertise of these domains for the evaluation. 
Evaluation on all source domains well performed as much as on the target domain as shown in Table~\ref{tab:exp_comparison_with_others_b}.
Notably, in PACS, A2XP achieved an average accuracy that was 2.9\% higher than the domain adaptation performance, as detailed in Table~\ref{tab:exp_different_inits}.

\subsection{Importance of Expert Processing}

Our study demonstrates that normalizing and scaling experts are crucial for the effective functioning of the A2XP module in mixing experts.
We conducted an ablation study focusing on three aspects: expert normalization, softmax, and the hyperbolic tangent function, with results detailed in Table~\ref{tab:ablation_study_exp_norm}. 
We calculated the performance gain of each factor by averaging the gain of every combination of the other two factors. 
Expert normalization makes experts initially have the same scales by following the normalization in Equation~\ref{eq:genralization_prompt}.
This normalization contributed to a significant accuracy gain of 39.09\% in the leave-one-domain-out evaluation.
The Softmax function takes a role as an amplifier of attention weights. 
It was observed to decrease the average accuracy by 4.35\%. 
This decrease is attributed to its tendency to significantly reduce the effect of experts with lower attention weights, even if the differences are insignificant.
The attention weights can be saturated during training since the calculation for each weight is independent of other experts. 
The Hyperbolic tangent function was applied to prevent such saturation problems and it led to 4.39\% accuracy gain. 
Consequently, the combination of expert normalization and hyperbolic tangent, without the softmax function, proved to be the most effective among the tested factor combinations.

\subsection{\track{Impact of Prompt Initialization}}

\begin{table}[t!]
    \centering \footnotesize
        \begin{tabular}{c|c|c|c}
        \toprule
        Expert Normalization & Softmax & $\tanh$ & Avg. Accuracy   \\
        \midrule
               &        &        & 49.35          \\
        \hline
        \cmark &        &        & 88.01          \\
               & \cmark &        & 46.96          \\
               &        & \cmark & 57.55          \\
        \midrule
               & \cmark & \cmark & 49.25          \\
        \cmark &        & \cmark & \textbf{95.07}          \\
        \cmark & \cmark &        & 88.19 \\
        \midrule
        \cmark & \cmark & \cmark & 88.19 \\
        \bottomrule
    \end{tabular}

    \vspace{-2mm}
    \caption{Ablation study about the A2XP module on PACS dataset.}
    \vspace{-2mm}
    \label{tab:ablation_study_exp_norm}
\end{table}

In this experiment, we compare several initialization strategies including zero, uniform distribution, Gaussian distribution, and meta prompting to justify the effectiveness of meta prompt initialization.
While good initialization might be optional for simpler tasks, its importance escalates with increasing task complexity.
For instance, as indicated in Table~\ref{tab:exp_different_inits}, meta prompt initialization did not show a marked advantage in scenarios where adaptation performance was not outstanding (see the result of PACS).
However, in more challenging situations, such as with the Office-Home dataset, meta prompt initialization significantly enhanced performance, from expert training through to generalization training. 
For example, the adaptation performance of zero initialization was the best among others which is 15.06\% lower accuracy. 
Consequently, the generalization performance of zero initialization is 11.21\% lower than meta prompt initialization. 
we set the number of updates to 10K for the evaluation of the Office-Home dataset.
It is noteworthy that there was no significant difference among other initialization strategies, and the correlation between adaptation and generalization was not linear. 
This suggests that effective expert adaptation is a critical foundation for A2XP, and good initialization is a key factor in achieving good adaptation.

\begin{table}[t!]
    \centering \footnotesize
        \begin{tabular}{c|ccccc}
        \toprule
                    & Picture & Art   & Cartoon & Sketch & Avg.  \\
        \midrule
        FT           & 23.71   & 42.72 & 56.61   & 29.12  & 38.04 \\
        LP          & 83.11   & 94.04 & 86.95   & 86.79  & 87.72 \\
        A2XP + FT   & 68.62   & 26.61 & 17.28   & 18.83  & 32.84 \\
        A2XP + LP   & \textbf{99.07}   & \textbf{95.27} & \textbf{98.07}   & \textbf{87.85}  & \textbf{95.07} \\
        \bottomrule
    \end{tabular}

    \vspace{-2mm}
    \caption{Comparison of tuning range on the objective network with and without A2XP. FT and LP refer to \textbf{F}ull \textbf{T}uning and \textbf{L}inear \textbf{P}robing, respectively.}
    \vspace{-3mm}
    \label{tab:exp_tuning_range}
\end{table}

\begin{figure}[t!]
\centering \fontsize{8}{9}
    \def\svgwidth{0.96\linewidth} \setlength\fboxrule{0pt}
    \fbox{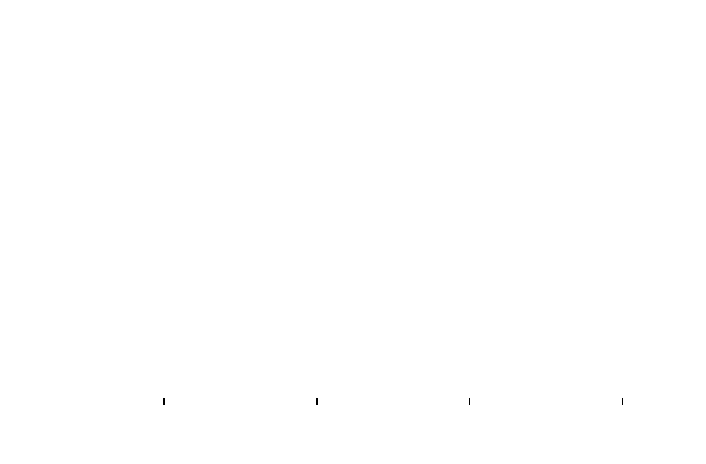}
    \caption{Comparison of two generalization strategies about fixing or tuning the experts in the generalization step.}
    \vspace{-3mm}
    \label{fig:exp_tune_more}
\end{figure}

\subsection{Effectiveness of A2XP Module}

We utilized a CLIP-pre-trained ViT as the objective network, which is also recognized for its well-generalized pre-trained model. 
We performed an ablation study to demonstrate the efficacy of the A2XP module by quantifying its impact on accuracy enhancement with commonly used fine-tuning approaches such as linear probing and full tuning. 
Initially, without A2XP, linear probing outperformed full tuning in the domain of generalization.
Specifically, linear probing achieved an average accuracy of 38.04\%, compared to 32.84\% for full tuning.
As shown in Table~\ref{tab:exp_tuning_range}, tuning the hidden layers appeared to impact the tuning of the output layer negatively.
With the integration of A2XP in linear probing, accuracy was significantly increased across all tested domains.
However, in the case of full tuning, the inclusion of A2XP was counterproductive.
We analyzed that full tuning is inherently unstable; thus, the A2XP module, positioned before the hidden layers, was adversely affected.
To summarize, further tuning might enhance average accuracy in certain scenarios, it generally leads to a decrease in accuracy and contributes to performance instability. 
Additionally, this implies that training experts through domain adaptation is more beneficial and effective compared to domain generalization.

\subsection{Further Expert Tuning}

\begin{figure}[t!]
    \centering \fontsize{8}{9}
    \def\svgwidth{0.98\linewidth} \setlength\fboxrule{0pt}
    \fbox{\hspace{-2mm} 
\begingroup%
  \makeatletter%
  \providecommand\color[2][]{%
    \errmessage{(Inkscape) Color is used for the text in Inkscape, but the package 'color.sty' is not loaded}%
    \renewcommand\color[2][]{}%
  }%
  \providecommand\transparent[1]{%
    \errmessage{(Inkscape) Transparency is used (non-zero) for the text in Inkscape, but the package 'transparent.sty' is not loaded}%
    \renewcommand\transparent[1]{}%
  }%
  \providecommand\rotatebox[2]{#2}%
  \newcommand*\fsize{\dimexpr\f@size pt\relax}%
  \newcommand*\lineheight[1]{\fontsize{\fsize}{#1\fsize}\selectfont}%
  \ifx\svgwidth\undefined%
    \setlength{\unitlength}{439.86455001bp}%
    \ifx\svgscale\undefined%
      \relax%
    \else%
      \setlength{\unitlength}{\unitlength * \real{\svgscale}}%
    \fi%
  \else%
    \setlength{\unitlength}{\svgwidth}%
  \fi%
  \global\let\svgwidth\undefined%
  \global\let\svgscale\undefined%
  \makeatother%
  \begin{picture}(1,0.8742329)%
    \lineheight{1}%
    \setlength\tabcolsep{0pt}%
    \put(0.23007122,0.84529459){\makebox(0,0)[lt]{\lineheight{1.25}\smash{\begin{tabular}[t]{l}Picture\end{tabular}}}}%
    \put(0.40423904,0.84529459){\makebox(0,0)[lt]{\lineheight{1.25}\smash{\begin{tabular}[t]{l}Art painting\end{tabular}}}}%
    \put(0.63667424,0.84529459){\makebox(0,0)[lt]{\lineheight{1.25}\smash{\begin{tabular}[t]{l}Cartoon\end{tabular}}}}%
    \put(0.8513307,0.84529459){\makebox(0,0)[lt]{\lineheight{1.25}\smash{\begin{tabular}[t]{l}Sketch\end{tabular}}}}%
    \put(0.00511561,0.70888903){\makebox(0,0)[lt]{\lineheight{1.25}\smash{\begin{tabular}[t]{l}(a) Original\end{tabular}}}}%
    \put(0.00511561,0.50428059){\makebox(0,0)[lt]{\lineheight{1.25}\smash{\begin{tabular}[t]{l}(b) Prompt\end{tabular}}}}%
    \put(0.00511561,0.29967215){\makebox(0,0)[lt]{\lineheight{1.25}\smash{\begin{tabular}[t]{l}(c) Gain\end{tabular}}}}%
    \put(0.00511561,0.0950637){\makebox(0,0)[lt]{\lineheight{1.25}\smash{\begin{tabular}[t]{l}(d) Loss\end{tabular}}}}%
    \put(0,0){\includegraphics[width=\unitlength,page=1]{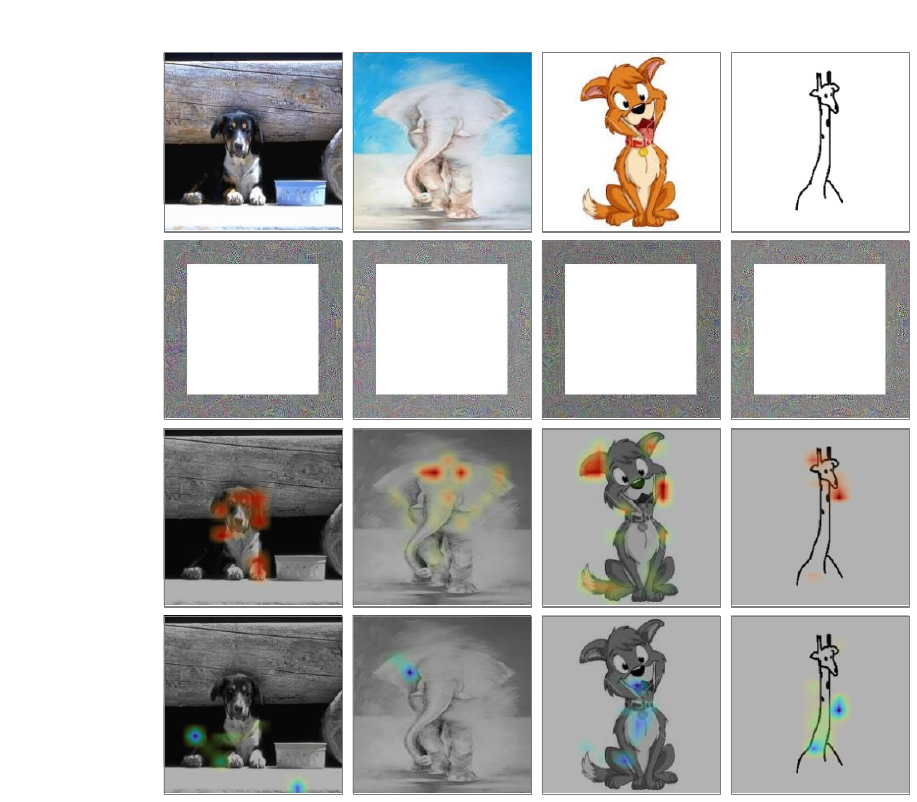}}%
  \end{picture}%
\endgroup%
}
    \caption{Activation visualization of A2XP using Grad-CAM~\cite{grad_cam}. (a) shows the input image, (c) and (d) show the relative gain and loss of activation using A2XP prompts in (b), respectively.}
    \vspace{-3mm}
    \label{fig:activation}
\end{figure}
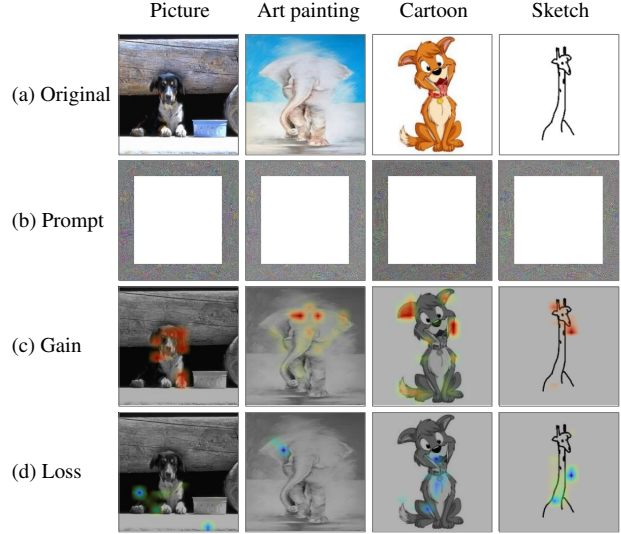

We carried out further experiments with a focus on generalization strategies, concentrating specifically on the experts rather than solely on the networks.
The premise was that further tuning of the experts during the generalization phase would facilitate the sharing of domain-specific knowledge among them.
To validate the effect of further tuning, we repeated the training ten times on PACS dataset, each time using a different fixed random seed.
The results of this experiment are depicted in Figure~\ref{fig:exp_tune_more}.
In the Picture domain, we observed a slight drop in mean accuracy, although this change was not statistically significant. The Art and Cartoon domains exhibited similar results, with average accuracies decreasing by 0.60\% and 1.60\%, respectively.
Notably, the standard deviation in both these domains increased significantly by 0.40\%. 
In contrast, the Sketch domain showed an improvement, with the average accuracy rising by 0.48\%, albeit accompanied by a similar increase in the standard deviation of 0.48\%.
This indicates that while further tuning of experts can lead to improvements in certain domains, it may also introduce greater variability in performance across different domains.

\begin{figure}[t]
    \centering \small
    \newcommand{\figsize}{0.48\linewidth}
    \newcommand{\narrowbottom}{\vspace{-3mm}}
    
    \begin{subfigure}{\figsize}
        \includegraphics[width=\linewidth]{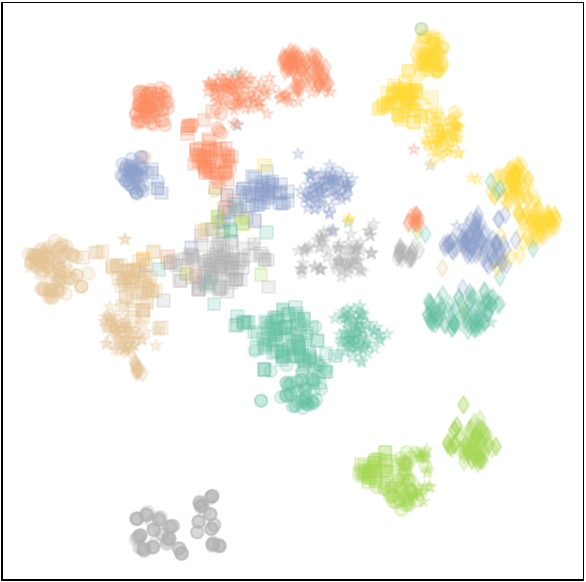}
        \caption{Picture}
        \narrowbottom
        \label{fig:manifold_a}
    \end{subfigure} \hspace{0.01\linewidth}
    \begin{subfigure}{\figsize}
        \includegraphics[width=\linewidth]{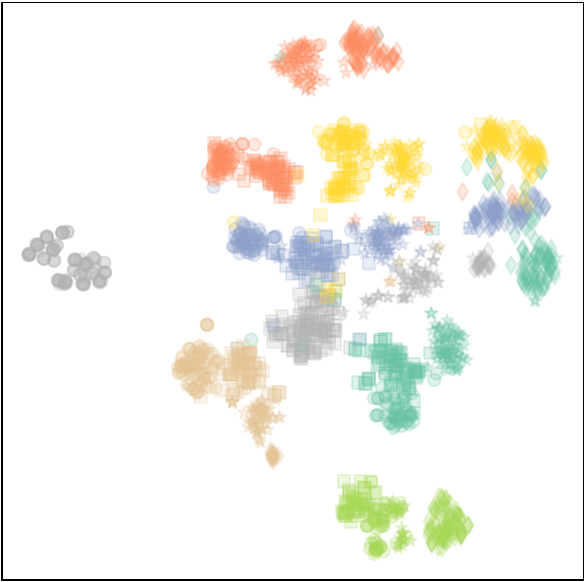}
        \caption{Art painting}
        \narrowbottom
        \label{fig:manifold_b}
    \end{subfigure} \hspace{0.01\linewidth}
    
    \begin{subfigure}{\figsize}
        \includegraphics[width=\linewidth]{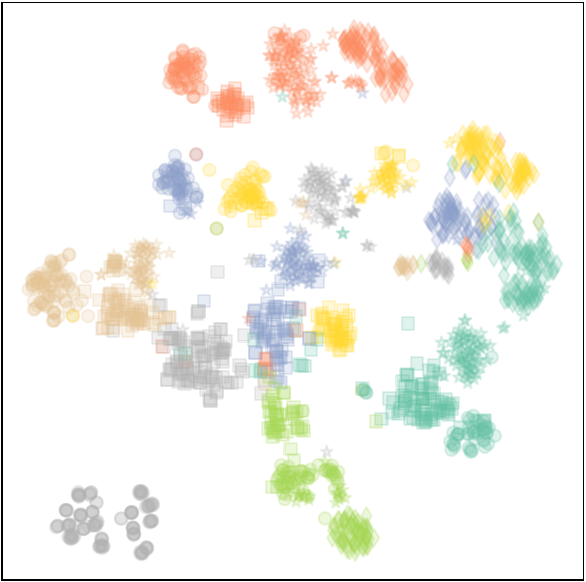}
        \caption{Cartoon}
        \narrowbottom
        \label{fig:manifold_c}
    \end{subfigure} \hspace{0.01\linewidth}
    \begin{subfigure}{\figsize}
        \includegraphics[width=\linewidth]{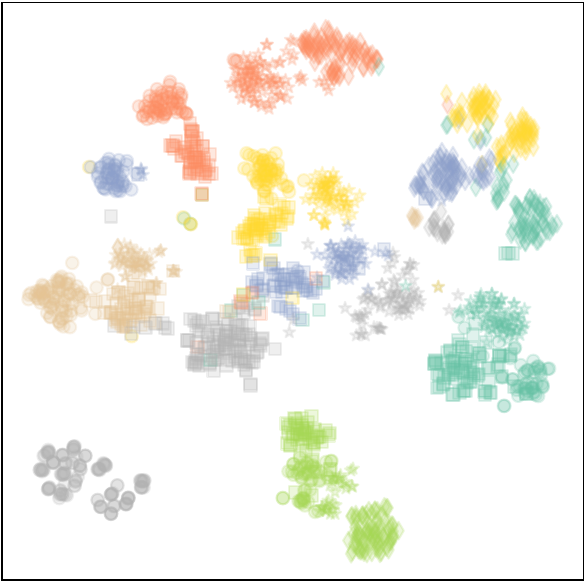}
        \caption{Sketch}
        \narrowbottom
        \label{fig:manifold_d}
    \end{subfigure} \hspace{0.01\linewidth}
    
    \begin{subfigure}{\figsize}
        \includegraphics[width=\linewidth]{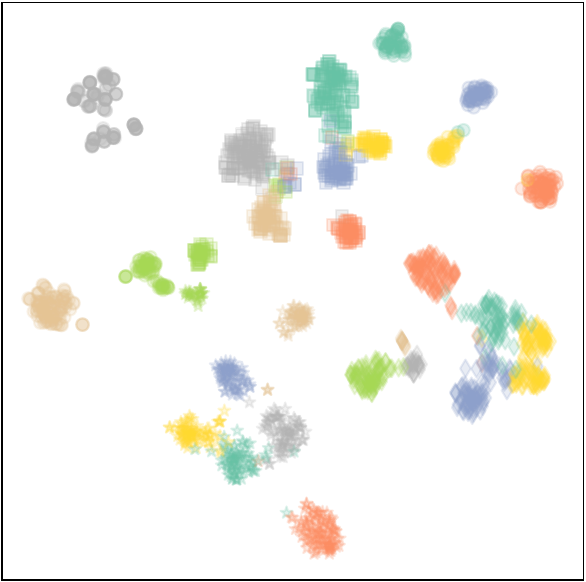}
        \caption{Adaptation (Experts)}
        \label{fig:manifold_e}
    \end{subfigure} \hspace{0.01\linewidth}
    \begin{subfigure}{\figsize}
        \centering \fontsize{6}{7}
        \def\svgwidth{0.9\linewidth} \setlength\fboxrule{0pt}
        \fbox{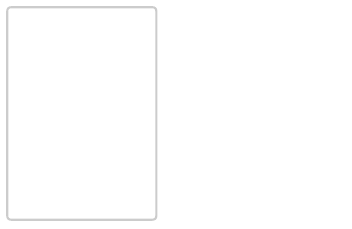}
        \vspace{12mm}
    \end{subfigure}

    \caption{t-SNE~\cite{tsne} visualization of correctly classified samples in manifold space. (a)-(d) illustrate the representation achieved through generalization, with Picture, Art Painting, Cartoon, and Sketch as the target domains. 
    (e) depicts the representation of expert adaptation prior to the generalization process.}
    \label{fig:manifold}
    \vspace{-3mm}
\end{figure}

\subsection{Visualization}

To help understand the effects of A2XP on the neural network's focus, we visualized the activation maps.
Table~\ref{tab:exp_tuning_range} demonstrates that while linear probing is generalized in a way, the takes the generalizability even further.
This suggests that linear probing without A2XP yields reasonably effective activation maps, and the incorporation of A2XP further refines and improves these activation maps.
Consequently, we extended our visualization beyond just the activation maps to include both the gains and losses in activation, as depicted in Figure~\ref{fig:activation}.
The prompts shown in the (b) row change the activation maps as much as shown in (c) and (d). 
The prompts have similar expression because they are from the same experts, but the intensities are different or some of them seem inverted. 
This means the experts are mixed in different ratios dependent on the target image. 
They show that A2XP makes the network attend more to the face representation and kills activation on other representations, such as the backgrounds or the body of an animal. 
Specifically in the Picture domain, (c) shows that it primarily activates the ears of the dog and deactivates the background. 
In the Sketch domain, it activates representations around the head while it deactivates the background next to the neck and the body, which contains fewer domain-agnostic clues for classification. 

Additionally, we visualized the manifold space of the features extracted from the last hidden layer, as shown in Figure~\ref{fig:manifold}, to observe how the classes and domains are represented in a 2-dimensional space.
Figure~\ref{fig:manifold_a}--\ref{fig:manifold_d} shows generalized features are mapped similarly regardless of the target domain.
Additionally, samples belonging to the same classes are closely grouped together, even when they originate from different domains. 
Conversely, as depicted in Figure~\ref{fig:manifold_e}, samples with the same class label but from different domains are mapped distinctly. 
It is understandable because the experts are trained independently, and the training does not concern other prompts to be mapped relevantly.

\subsection{Space Complexity Analysis}
We calculated the space complexity of A2XP compared to DART~\cite{dart}.
DART requires memory proportional to the number of augmentation presets $(M)$ while A2XP requires much less memory space with $N$ expert prompts.
Let the number of parameters of the objective network as $S_\symnet$, the big-$\mathbf{O}$ notation of DART and A2XP are
\begin{align}
    \mathbf{O}_\textnormal{DART}(M) &= MS_\symnet, \\
    \mathbf{O}_\textnormal{A2XP}(N) &= NS_\symprompt + S_\symnet + S_\mathcal{E} = NS_\symprompt,
    \label{eq:big-o}
\end{align}
where $S_\symprompt$ and $S_\mathcal{E}$ denote the number of parameters in a single prompt and the encoders, respectively. 
This demonstrates a key advantage of our method: its reduced memory usage compared to comparing approaches.

\section{Conclusion and Future Works}
\label{sec:conclusion}

In this work, we proposed a novel domain generalization method A2XP.
A2XP solves the domain generalization problem as a direction regression problem by disentangling it into two steps: domain adaptation and domain generalization. 
In the domain adaptation step, experts are trained on each source domain to take the place of a hint. 
In the domain generalization step, a network is trained to mix those experts properly depending on the target images. 
A2XP does not require changing the architecture or parameters of the objective network, which is the key to keeping the network private. 
A2XP outperformed state-of-the-art with a limited number of updates in PACS, VLCS datasets and successfully performed not only on the target domain but also on the source domains. 
We proved this problem definition mathematically based on the likelihood maximization problem. 
We also justified the effectiveness and characteristics by conducting extensive experimentation. 

Our work introduced a remarkable issue of privacy in domain generalization and proposed a powerful domain generalization method, but it also has limitations. 
A2XP requires well-trained experts for the domain generalization step. 
However, to the best of our knowledge, some datasets are difficult to adapt with input prompts. 
And the problems with adaptation techniques must be improved for A2XP to be widely used. 
We hope that this work encourages more research to solve this issue and improve this novel framework, and this will also be left as our future work.


\vspace{0.5\baselineskip}\noindent\textbf{Acknowledgement.} This work was partly supported by a National Research Foundation of Korea (NRF) grant funded by the Korean government (MSIT) (NRF-2022R1C1C1008074), and by an Institute of Information and Communications Technology Planning and Evaluation (IITP) grant funded by the Korean government (MSIT) (No.RS-2022-00155911, Artificial Intelligence Convergence Innovation Human Resources Development (Kyung Hee University)).

{
    \small
    \bibliographystyle{ieeenat_fullname}
    \bibliography{main}
}


\end{document}


\maketitlesupplementary

\section{Implementation of Generalization}
\label{sec:implementation_detail}

In this section, we present detailed implementation of the \textit{Attention-based Generalization} module in a pseudo-code form from initialization to forwarding~Algorithm~\ref{alg:a2xp_pytorch}. 

\begin{algorithm}
\caption{Generalization Implementation}
\label{alg:a2xp_pytorch}
\newcommand{\self}{\textnormal{self}}
\newcommand{\symembed}{\mathbf{z}}

\begin{algorithmic}[1]
    \small
    \Procedure{init}{$\self, \symprompt_1, \symprompt_2, \cdots, \symprompt_i, \cdots, \symprompt_N$}
        \State $\self.\mathcal{E}_\textnormal{shared} \gets \operatorname{resnet18\_1k}()$
        \Comment{Initialize embedders.}
        \State $\self.\mathcal{E}_\textnormal{T}, \self.\mathcal{E}_\textnormal{E} \gets \operatorname{linear}(), \operatorname{linear}()$
        \State $\self.\symprompt_i \gets \symprompt_i / \Vert \symprompt_i \Vert_2 \quad \forall{i \in [1,N]}$
        \Comment{Normalize experts.}
    \EndProcedure \vspace{3mm}

    \Procedure{forward}{$\self, \syminput_{N+1,j}$}
        \State $\symembed_\syminput \gets \self.\mathcal{E}_\textnormal{T}(\self.\mathcal{E}_\textnormal{shared}(\syminput_{N+1,j}))$
        \State $\symembed_{\symprompt_i} \gets \self.\mathcal{E}_\textnormal{E}(\self.\mathcal{E}_\textnormal{shared}(\self.\symprompt_i)) \quad \forall{i \in [1,N]}$

        \State $\lambda_i \gets \symembed_\syminput \symembed_{\symprompt_i}^\top \quad \forall{i \in [1,N]}$
        \Comment{Calculate attention scores.}
        \State $\symprompt_{N+1,j} \gets \sum_{i=1}^{N}{\lambda_i \self.\symprompt_i}$
        \State \textbf{return} $\syminput_{N+1,j} + \symprompt_{N+1,j}$
    \EndProcedure
\end{algorithmic}
\end{algorithm}
\vspace{-3mm}

\section{Further Analysis on the Experts}
\label{sec:analysis_experts}

In this section, we conduct further analysis on the expert prompts of A2XP. 
We analyzed the prompts by changing various components: the size of the experts, the number of experts, the type of prompts, the way to mix the experts.

\subsection{Size of the Experts}

\begin{figure}[h]
    \centering
    \includegraphics[width=\linewidth]{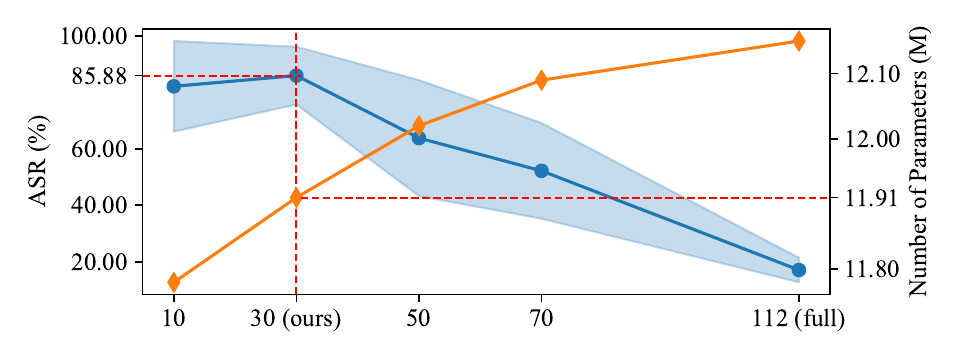}
    \vspace{-8mm}
    \caption{Expert adaptation performance of A2XP with Gaussian initialization. The blue transparent range shows $\mu \pm \sigma$ of ASR.}
    \label{fig:prompt_size}
\end{figure}

We analyzed the prompt size in the performance and the memory requirement perspectives (see Figure~\ref{fig:prompt_size}).
We empirically found that 30 is the best prompt size among the five sizes and applied it to our method.

\vspace{-3mm}
\vspace{-4.5mm}

\subsection{Ablation Study on Various Experts}

We compared domain generalization performance among various experts. 

\subsubsection{Various Prompts}

Generalization \textit{without} prompts, which is equivalent to linear probing, loses the benefits of linear combination; therefore, it has a lower generalization performance (see Table~\ref{tab:prompt_kind}).  
Utilizing \textit{random} prompts show performance improvement, indicating that prompts can contribute to better generalization performance. 
Furthermore, we show the effectiveness of our method that enhances generalization performance more efficiently by leveraging \textit{experts} trained from each domain.

\begin{table}[h]
    \centering \footnotesize
    \begin{tabular}{c|ccccc}
        \toprule
                        & Picture & Art   & Cartoon & Sketch & Avg.  \\
        \midrule
        Without         & 86.95   & 83.11 & 94.04   & 86.79  & 87.72 (-7.35) \\
        Random          & 98.98   & 93.85 & 90.19   & 88.09  & 92.78 (-2.29) \\
        \rowcolor{Gray} 
        Experts         & \textbf{99.07}   & \textbf{95.27} & \textbf{98.07}   & \textbf{87.85}  & \textbf{95.07} (-0.00) \\
        \bottomrule
    \end{tabular}
    \vspace{-1mm}
    \caption{Comparison by various prompts.}
    \label{tab:prompt_kind}
\end{table}
\vspace{-6mm}

\subsubsection{Number of Experts}

As shown in Table~\ref{tab:num-experts}, using either no experts or just a single expert offers limited generalization potential.
In contrast, employing multiple experts, particularly in numbers matching the domain count, broadens the scope to identify the optimal direction for generalization.
\begin{table}[h]
    \centering \footnotesize
    \begin{tabular}{l|ccccc}
        \toprule
        \# experts  & Picture & Art   & Cartoon & Sketch & Avg.  \\
        \midrule
        0           & 86.95   & 83.11 & 94.04   & 86.79  & 87.72 (-7.35) \\
        1           & 83.75   & 95.69 & 86.82   & 86.49  & 88.19 (-6.88) \\
        2           & 97.35   & \textbf{99.34} & 93.41   & 87.27  & 94.34 (-0.73) \\
        \rowcolor{Gray} 
        3 (all)     & \textbf{99.07}   & 95.27 & \textbf{98.07}   & \textbf{87.85}  & \textbf{95.07} (-0.00) \\
        \bottomrule
    \end{tabular}
    \vspace{-1mm}
    \caption{Comparison by \# experts.} 
    \label{tab:num-experts}
\end{table}
\vspace{-3mm}

\subsection{Various Ways to Mix Experts}

In Table~\ref{tab:mix-experts}, we compared various methods to mix pre-trained experts for unseen datasets.
We demonstrate that our \textit{attention}-based approach outperforms methods that mix experts in \textit{constant} or \textit{random} weights. 

\begin{table}[h]
    \centering \footnotesize
    \begin{tabular}{c|ccccc}
        \toprule
                    & Picture & Art   & Cartoon & Sketch & Avg.  \\
        \midrule
        Constant    & 97.82   & 99.40 & 94.24   & 86.99  & 94.61 (-0.46) \\
        Random      & 97.65   & 99.16 & 93.85   & 87.05  & 94.43 (-0.64) \\
        \rowcolor{Gray}
        Attention   & \textbf{99.07}   & \textbf{95.27} & \textbf{98.07}   & \textbf{87.85}  & \textbf{95.07} (-0.00) \\
        \bottomrule
    \end{tabular}
    \vspace{-1mm}
    \caption{Various ways to mix experts.}
    \label{tab:mix-experts}
\end{table}
\vspace{-3mm}

\section{Further Analysis on the Framework}
\label{sec:analysis_a2xp}

We analyzed more about the A2XP framework itself in the perspective of how evenly the experts are mixed, how does the objective network architecture affects, and the scalability of A2XP. 

\subsection{Attention Distribution}

\begin{figure*}[ht]
    \centering \small
    
    \begin{subfigure}{\linewidth}
        \centering
        \includegraphics[width=0.95\linewidth]{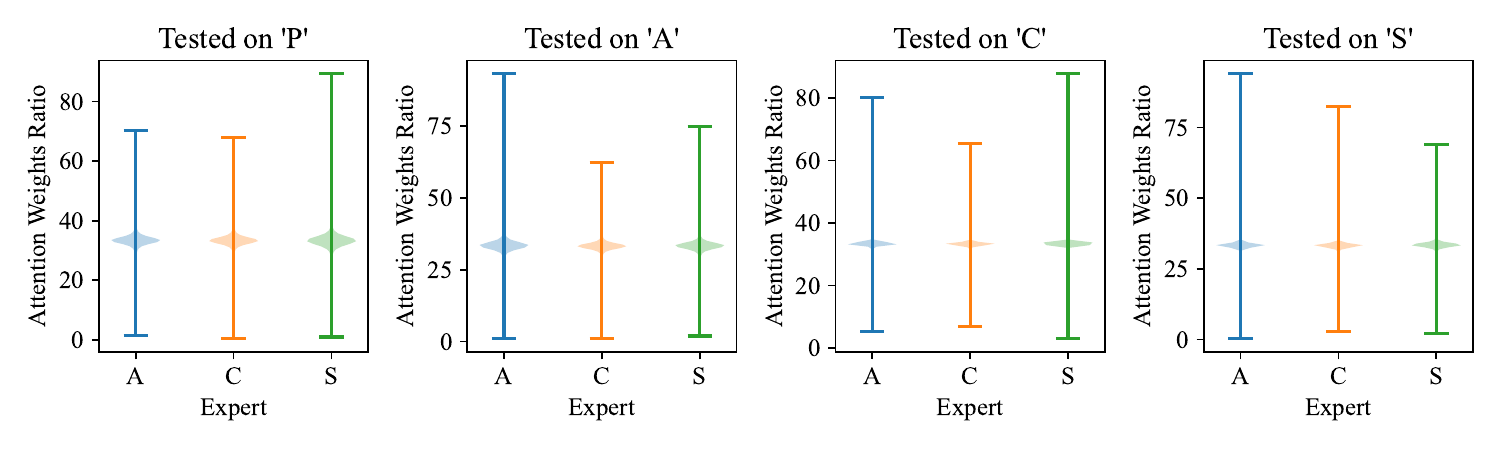}
        \vspace{-2mm}
        \caption{Trained on `P'}
        \vspace{-1mm}
        \label{fig:attn_weights_p}
    \end{subfigure}
    \begin{subfigure}{\linewidth}
        \centering
        \includegraphics[width=0.95\linewidth]{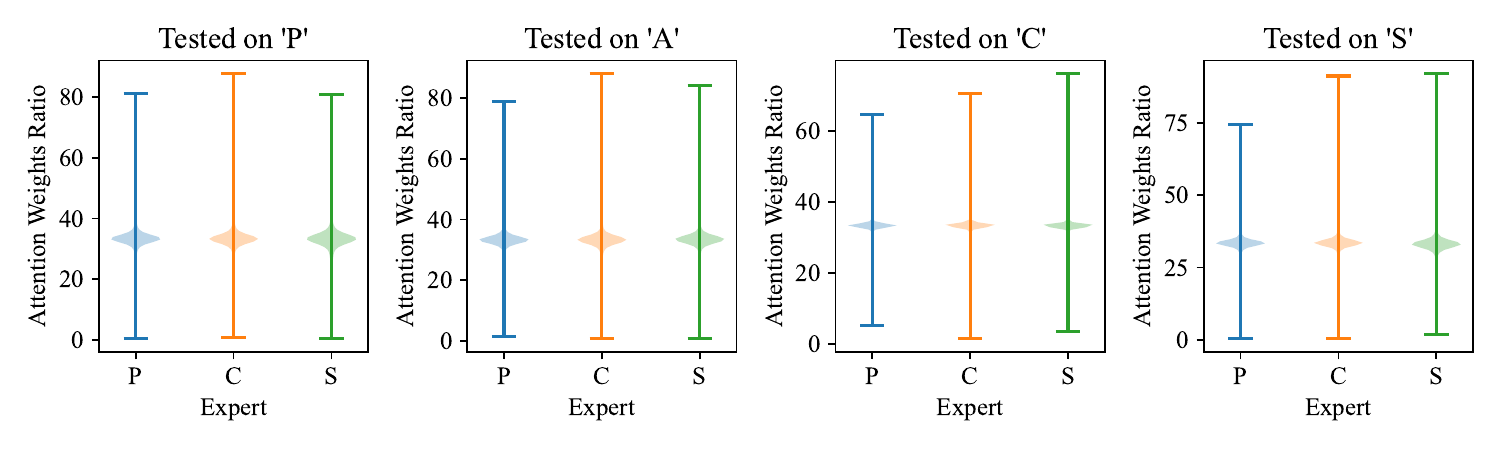}
        \vspace{-2mm}
        \caption{Trained on `A'}
        \vspace{-1mm}
        \label{fig:attn_weights_a}
    \end{subfigure}
    \begin{subfigure}{\linewidth}
        \centering
        \includegraphics[width=0.95\linewidth]{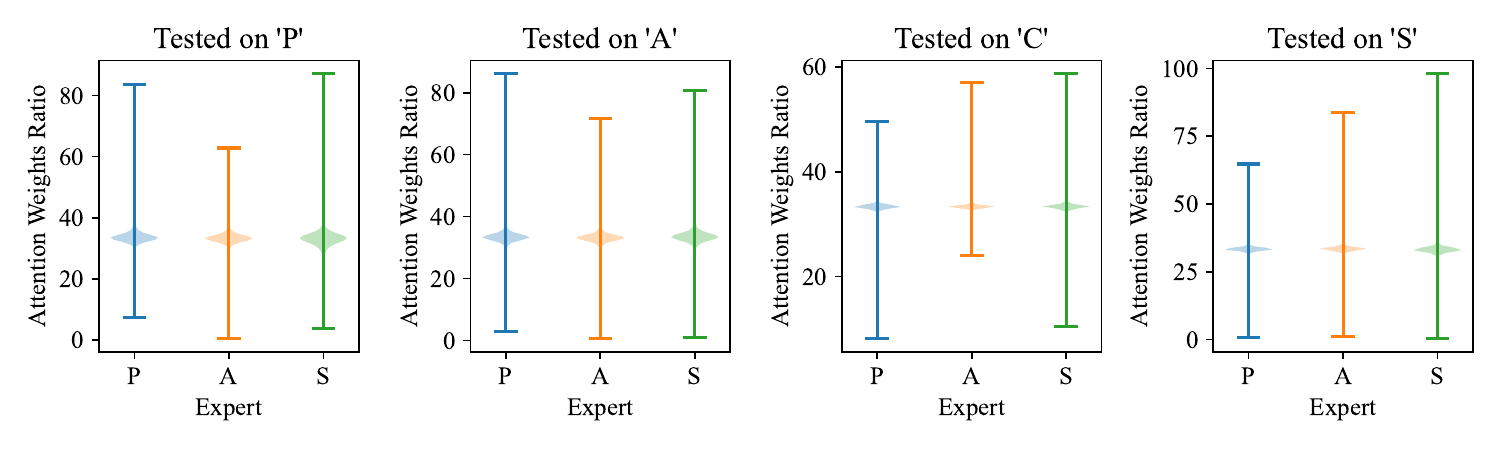}
        \vspace{-2mm}
        \caption{Trained on `C'}
        \vspace{-1mm}
        \label{fig:attn_weights_c}
    \end{subfigure}
    \begin{subfigure}{\linewidth}
        \centering
        \includegraphics[width=0.95\linewidth]{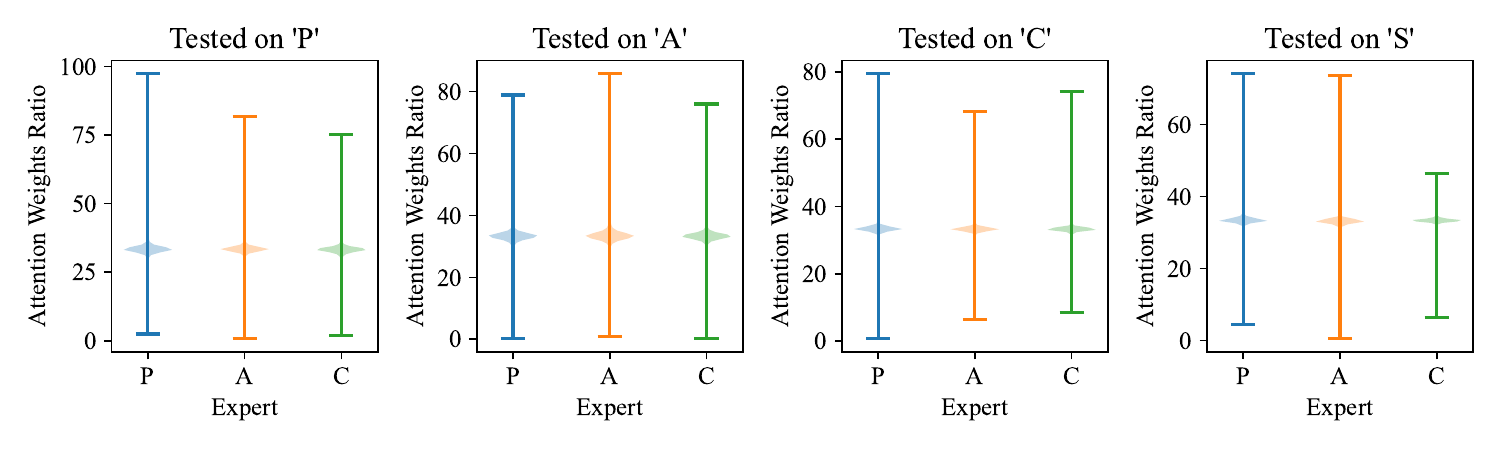}
        \vspace{-2mm}
        \caption{Trained on `S'}
        \vspace{-1mm}
        \label{fig:attn_weights_s}
    \end{subfigure}

    \vspace{-1mm}
    \caption{Visualization of normalized attention weights of correctly classified samples from A2XP on PACS~\cite{pacs} dataset.}
    \label{fig:attn_weights}
\end{figure*}

When A2XP is applied on the source domain, we expected the attention weights of A2XP emphasize the experts of the source domain.
This study analyzes how A2XP attends to different experts depending on the domain of the input images.
The violin plots in Figure~\ref{fig:attn_weights} show the distribution of normalized attention weights in PACS~\cite{pacs} dataset.
Each cell shows the distribution of attention weights on each domain. 
Across all combinations of target and source domains, a significant standard deviation was observed, indicating a wide range of variation in the attention weights.
This suggests that the attention weights have a very large range.
\begin{table}[h]
    \centering \footnotesize
    \begin{tabular}{c|cccc}
         \toprule
           & P & A & C & S \\
         \midrule
         P & 1.729E-1 & \textbf{1.330E-2} & 3.424E-1 & \textbf{2.377E-4} \\
         A & 4.966E-1 & 5.752E-2 & \textbf{4.210E-2} & 5.739E-2 \\
         C & \textbf{2.127E-2} & \textbf{1.641E-3} & 1.759E-1 & \textbf{1.797E-2} \\
         S & 2.556E-1 & 2.526E-1 & 5.566E-1 & \textbf{2.460E-9} \\
         \bottomrule
    \end{tabular}
    \caption{$p$-values of RM-ANOVA~\cite{rm_anova} with the normalized attention weights on PACS~\cite{pacs} dataset. Bold styled cells are significant with $p \le 0.05$.}
    \label{tab:rm_anova}
\end{table}

To be analytic, we performed Repeated Measures-ANalysis Of VAriance (RM-ANOVA)~\cite{rm_anova} on the normalized attention weights, and the result is in Table~\ref{tab:rm_anova}.
Each cell contains the $p$-value of a combination of the target domain and tested domain. 
For example, $p$-value of weights when trained on `P' and tested on `A' is 1.330E-2. 
In this case, the experts are from the `A,' `C,' and `S' domains.
The smaller a $p$-value is, the more the combination showed a significant correlation among weights for experts. 
The $p$-values are significant with $p \le 0.05$ in some cases but not dominant. 
As a result, A2XP mixes the experts differently depending to the input images, and the mixing ratios are not always similar even if the target and testing domain is the same. 

\subsection{Various Objective Networks}

We are concerned only about CLIP~\cite{clip}-pretrained Vision Transform (ViT)~\cite{vit} for the objective network in the main paper. 
We present another result on a convolutional neural network ResNet50~\cite{resnet} and ImageNet~\cite{imagenet} supervised pretraining to reveal another characteristic of A2XP.
The leave-one-domain-out evaluation result is compared in Table~\ref{tab:resnet50}.
The number of updates was limited to 3K for ImageNet and 1K for CLIP pretrained models in the adaptation step.
And we initialized the experts by zero before adaptation.

We observed that the experts must be well adapted for all domain from ResNet50 with both ImageNet and CLIP pretraining.
Moreover, even if the adaptation was successful, the model itself have to be generalized at the pretext task. 
Both the average accuracy of the both ResNet50 was lower compared to other existing methods~\cite{miro,dart}. 
As a result, A2XP is sensitive to the adaptation method, the objective network architecture, and the pretext task. 

\begin{table}[h]
    \centering \footnotesize
    \renewcommand{\tabcolsep}{1mm}
    \begin{tabular}{cc|ccccc}
        \toprule
                               &                          & \multicolumn{5}{c}{Expert Adaptation}              \\
        Architecture           & Pretraining              & P       & A            & C       & S      & Avg.   \\
        \midrule
        ResNet50~\cite{resnet} & ImageNet~\cite{imagenet} & 92.40   & 72.36        & 85.24   & 66.28  & 79.07  \\
        ResNet50~\cite{resnet} & CLIP~\cite{clip}         & 67.25   & 52.83        & 59.98   & 56.73  & 59.20  \\
        ViT-base~\cite{vit}    & ImageNet~\cite{imagenet} & 96.95   & 79.30        & 92.41   & 87.94  & 89.15  \\
        \rowcolor{Gray}
        ViT-base~\cite{vit}    & CLIP~\cite{clip}         & 97.54   & 73.88        & 95.52   & 94.55  & 90.37  \\
        \midrule \midrule
                               &                          & \multicolumn{5}{c}{Attention-based Generalization} \\
        Architecture           & Pretraining              & P       & A            & C       & S      & Avg.   \\
        \midrule
        ResNet50~\cite{resnet} & ImageNet~\cite{imagenet} & 51.56   & 49.12        & 46.25   & 36.12  & 45.76  \\
        ResNet50~\cite{resnet} & CLIP~\cite{clip}         & 74.31   & 44.38        & 42.62   & 16.34  & 44.41  \\
        ViT-base~\cite{vit}    & ImageNet~\cite{imagenet} & 81.02   & 69.53        & 49.23   & 31.38  & 57.79  \\
        \rowcolor{Gray}
        ViT-base~\cite{vit}    & CLIP~\cite{clip}         & 99.07   & 95.07        & 98.12   & 88.22  & 95.12  \\
        \bottomrule
    \end{tabular}
    
    \caption{The result of leave-one-domain-out evaluation using ViT~\cite{vit} and ResNet50~\cite{resnet}.}
    \label{tab:resnet50}
\end{table}
\vspace{-3mm}

\subsection{Scalability}

We applied our method to larger datasets: Office-Home~(Table~\ref{tab:officehome-eval}) and DomainNet~(Table~\ref{tab:domainnet-eval}). 
The results show that A2XP outperforms current methods, validating its applicability across datasets of varying sizes.

\begin{table}[h]
    \centering \footnotesize \renewcommand{\tabcolsep}{1.4mm} \renewcommand{\arraystretch}{1}
    \begin{tabular}{c|ccccc}
        \toprule
                                    & Art     & Clipart     & Product     & Real     & Avg.  \\
        \midrule
        ERM                         & 48.04 & 42.27 & 48.25 & 47.63 & 46.55 \\
        MIRO                        & 56.49 & 58.56 & 43.30 & 54.43 & 53.20 \\
        \rowcolor{Gray}
        A2XP                        & \textbf{77.42} & \textbf{65.73} & \textbf{81.93} & \textbf{83.15} & \textbf{77.06} \\
        \bottomrule
    \end{tabular}
    \vspace{-1mm}
    \caption{Office-Home evaluation.}
    \vspace{-4mm}
    \label{tab:officehome-eval}
\end{table}
\begin{table}[h]
    \newcommand{\stickybottom}{\vspace{-4mm}}
    \centering \footnotesize \renewcommand{\tabcolsep}{1.4mm} \renewcommand{\arraystretch}{1}
    \begin{tabular}{c|ccccccc}
        \toprule
                                    & Clip  & Info  & Paint & Quick & Real  & Sketch & Avg.  \\
        \midrule
        ERM                         & 0.32  & 0.35  & 0.45  & 0.39  & 0.41  & 0.57   & 0.41  \\
        MIRO                        & 39.31 & 39.48 & 40.10 & \textbf{39.77} & 40.59 & 42.18  & 40.24 \\
        \rowcolor{Gray}
        A2XP                        & \textbf{62.88} & \textbf{43.58} & \textbf{58.99} & 13.72 & \textbf{55.96} & \textbf{58.45}  & \textbf{48.93} \\
        \bottomrule
    \end{tabular}
    \vspace{-1mm}
    \caption{DomainNet evaluation.}
    \label{tab:domainnet-eval}
\end{table}

We further investigated the scalability of A2XP for datasets of various sizes by measuring the number of parameters, memory requirements, computational resources~(GFLOPs), and the training time (see Table~\ref{tab:scalability}).
The number of parameters and the memory requirement of A2XP only depends on the number of experts. 
Training time primarily depends on the number of training samples.
From this perspective, we show the practical applicability of A2XP for larger datasets. 

\begin{table}[h]
    \centering \tiny \renewcommand{\tabcolsep}{1mm} \renewcommand{\arraystretch}{0.8}
    \begin{tabular}{c|ccc|ccc|cc}
        \toprule
        Dataset     & \# classes & \# domains & \# samples & \# params & Mem. load & GFLOPs   & Time (s) & Avg. Acc.  \\
        \midrule
        PACS        & 7          & 4          & 9,991      & 11.91M    & 17817MiB  & 1.814    & 2.12     & 95.07 \\
        VLCS        & 5          & 4          & 10,729     & 11.91M    & 17777MiB  & 1.814    & 2.51     & 83.15 \\
        Office-Home & 65         & 4          & 15,588     & 11.91M    & 17779MiB  & 1.814    & 2.87     & 77.06 \\
        DomainNet   & 345        & 6          & 586,575    & 12.05M    & 18185MiB  & 2.539    & 136.46   & 48.93 \\
        \bottomrule
    \end{tabular}
    \vspace{-1mm}
    \caption{Scalability analysis.}
    \vspace{-2mm}
    \label{tab:scalability}
\end{table}

\section{Discussion}

\subsection{Failure Cases}

We found that A2XP struggles to generalize for specific domains such as \textit{Quick} in the DomainNet dataset and \textit{Sketch} domain in the PACS dataset; both have more significant domain shifts from another dataset. 
Despite limitations in generalizing distinct domains, the performance of our approach still achieved state-of-the-art average accuracy. 

\subsection{Interpretability}

As noted in~\cite{vpt}, a visual prompt facilitates domain adaptation by aligning features between the source and target domains.
This can be interpreted as our experts are responsible for shifting features towards the target domain's manifolds. 
In our privacy setting, the alignment target is the manifold characterized by features from the data used to pre-train the model. 
Consequently, generating an expert for an unseen domain by mixing the experts from other domains can be considered crafting a mapping function to the pre-trained manifold, which we interpret as contributing to enhancing decision-making when using a pre-trained model while keeping it private.


{
    \small
    \bibliographystyle{ieeenat_fullname}
    \bibliography{main}
}